# Assessing The Performance of YOLOv5 Algorithm For Detecting Volunteer Cotton Plants in Corn Fields at Three Different Growth Stages

**Pappu Kumar Yadav**
Texas A&M University
pappuyadav@tamu.edu

**J. Alex Thomasson**
Mississippi State University
athomasson@abe.msstate.edu

**Stephen W. Searcy**
Texas A&M University
s-searcy@tamu.edu

**Robert G. Hardin**
Texas A&M University
rghardin@tamu.edu

**Ulisses Braga-Neto**
Texas A&M University
ulisses@tamu.edu

**Sorin C. Popescu**
Texas A&M University
s-popescu@tamu.edu

**Daniel E. Martin**
U.S.D.A. Agriculture Research Service
dan.martin@usda.gov

**Roberto Rodriguez**
U.S.D.A.- APHIS PPQ S&T
Roberto.RodriguezIII@usda.gov

**Karem Meza**
Utah State University
karem.mezacapcha@usu.edu

**Juan Enciso**
Texas A&M AgriLife Research
And Extension Center at Weslaco
Juan.Enciso@ag.tamu.edu

**Jorge Solorzano Diaz**
Texas A&M AgriLife Research
And Extension Center at Weslaco
Jorge.SolorzanoDiaz@ag.tamu.edu

**Tianyi Wang**
China Agriculture University
tianyi.wang@cau.edu.cn

## Abstract

The boll weevil (*Anthonomus grandis* L.) is a serious pest that primarily feeds on cotton plants. In places like Lower Rio Grande Valley of Texas, due to sub-tropical climatic conditions, cotton plants can grow year-round and therefore the left-over seeds from the previous season during harvest can continue to grow in the middle of rotation crops like corn *(Zea mays* L.) and sorghum (*Sorghum bicolor* L.). These feral or volunteer cotton (VC) plants when reach the pinhead squaring phase (5-6 leaf stage) can act as hosts for the boll weevil pest. The Texas Boll Weevil Eradication Program (TBWEP) employs people to locate and eliminate VC plants growing by the side of roads or fields with rotation crops but the ones growing in the middle of fields remain undetected. In this paper, we demonstrate the application of computer vision (CV) algorithm based on You Only Look Once version 5 (YOLOv5) for detecting VC plants growing in the middle of corn fields at three different growth stages (V3, V6 and VT) using unmanned aircraft systems (UAS) remote sensing imagery. All the four variants of YOLOv5 (s, m, l, and x) were used and their performances were compared based on classification accuracy, mean average precision (mAP) and F1-score. It was found that YOLOv5s could detect VC plants with maximum classification accuracy of 98% and mAP of 96.3 % at V6 stage of corn while YOLOv5s and YOLOv5m resulted in the lowest classification accuracy of 85% and YOLOv5m and YOLOv5l had the least mAP of 86.5% at VT stage on images of size 416 x 416 pixels. The developed CV algorithm has the potential to effectively detect and locate VC plants growing in the middle of corn fields as well as expedite the management aspects of TBWEP.

# 1. Introduction

The boll weevil (*Anthonomus grandis* L.) is a serious pest that primarily feeds on cotton plants and has cost the U.S. cotton industry more than 23 billion USD in economic losses since it first entered to the U.S. from Mexico in the 1890s (Harden, 2018). The National Boll Weevil Eradication Program (NBWEP) has successfully eradicated the boll weevils from major parts of the U.S.; however southern parts of Texas (the Lower Rio Grande Valley) remain prone to re-infestation each year due to its sub-tropical climatic conditions and proximity to the Mexico border (Roming et al., 2021). The sub-tropical climatic conditions allow cotton plants to grow year-round and therefore the left-over cotton seeds during the harvest from previous season can grow either at the edges of fields or in the middle of rotation crops like corn (*Zea mays* L.) and sorghum (*Sorghum bicolor* L.) (Wang et al., 2022; Yadav et al., 2019; Yadav et al., 2022). These feral or volunteer cotton (VC) plants when reach the pinhead squaring phase (5-6 leaves) can serve as hosts for the boll weevil pests (Yadav et al., 2022).

As per the management practices of The Texas Boll Weevil Eradication Foundation (TBWEF), edges of rotation crop fields are inspected for the presence of VC plants and are eliminated if found, to avoid pest re-infestation in future. However, VC plants growing in the middle of fields remain undetected as it is practically not possible to detect and locate them in thousands of acres of fields. Therefore, if any boll weevil is found to be trapped in the pheromone traps of such fields, the whole field is sprayed with chemicals like Malathion ULV (FYFANON® ULV AG) to kill the pests (National Cotton Council of America, 2012). Usually spray capable unmanned aircraft systems (UAS) are used for this purpose and the typical spray rate ranges from 0.56 to 1.12 kg/ha (FMC Corporation, 2001). Malathion is classified as toxicity class III pesticide and therefore can be dangerous to humans when not applied wisely (United States Environmental Protection Agency, 2016). Apart from this, excessive spraying of the chemical may kill beneficial insects and pests in the fields of rotation crops. Therefore, it has become a necessity to detect and locate VC plants growing in the middle of rotation crops. This will have two-fold advantages: the first one is that once the VC plants in the middle of fields are detected and located, spray capable UAS can spot-spray herbicides to eliminate the VC plants before they reach the pinhead squaring phase, secondly, if some of them survive and grow beyond the pinhead squaring phase due to herbicide tolerance or resistance then Malathion can be spot-sprayed at the detected VC plants instead of spraying the entire field due to the fact that boll weevil pests are most likely to be found at the VC plants as they feed on them. Hence, detecting the VC plants is crucial for the management aspects of TBWEF because it can speed up the management practices as well as help minimize the chemical costs associated with herbicides and pesticides.

Detecting VC plants growing in the middle of corn or sorghum fields is a two-step process i.e., classification followed by localization which necessarily means object detection task (Howard et al., 2019; Mustafa et al., 2020; Silwal et al., 2021; Zhao et al., 2019). In our past study, pixel-wise classification method was used with classical machine learning algorithms to detect the regions of VC plants in a corn field (Yadav et al., 2019). This approach resulted in a classification accuracy of around 70% which was far below our expectations and therefore couldn't be used for practical applications. In another approach by Westbrook et al. (2016), traditional image processing method-linear spectral unmixing was used to detect individual cotton plants at an early growth stage with fairly good accuracy. In recent past, many deep learning-based algorithms have been developed and used successfully for object detection tasks. Many of those algorithms use convolution neural networks (CNNs) some of which are YOLOv3 (Redmon & Farhadi, 2018), YOLOv5 (Jocher, 2020), Mask R-CNN (He et al., 2017), etc. In this paper, we have shown the application of YOLOv5 for detecting VC plants growing in the middle of corn fields by using remote sensing multispectral imagery collected by unmanned aircraft systems (UAS).YOLOv5 is a one stage object detection algorithm that was originally released in four different variants (YOLOv5s, YOLOv5m, YOLOv5l and YOLOv5x) based on the network depth and number of parameters (Jocher, 2020; Jocher et al., 2021). The letters *s, m, l,* and *x* represent small, medium, large, and extra-large respectively depending upon the network depth and parameter size used. The specific objective of this paper is to do a comparative analysis of the performances of all the four variants of YOLOv5 for detecting VC plants in corn fields at three different growth stages (V3, V6 and VT) of corn plants. This study is an extension of our previous studies in which both YOLOv3 and YOLOv5 were used successfully in detecting VC plants at a single growth stage of corn plants (Yadav et al., 2022; Yadav et al., 2022).

## 2. Materials and methods

### 2.1 Experiment sites

The experiment sites were located at two different corn fields: one was based in Hidalgo county near Weslaco, Texas (97°56'21.538"W, 26°9'49.049"N) while the other was based in Burleson county near College Station, Texas (96°25'45.9"W, 30°32'07.4"N) (Figure 1). Two types of soil (Harlingen clay and Raymondville clay loam) are primarily present at the experiment plot based in Weslaco, TX while Weswood silty clay loam, Yahola fine sandy loam and Belk clay are present at the experiment site of Burleson county (USDA-Natural Resources Conservation Service, 2020). To mimic the presence of VC plants, 105 cotton seeds of variety Phytogen 350 W3FE (CORTEVA agriscience, Wilmington, Delaware) were planted in the corn field of Weslaco. Some of these were planted in line with corn plants while the rest were planted in the furrow middles. Similarly, a total of 180 cotton seeds-90 of the variety Phytogen 340 W3FE (CORTEVA agriscience, Wilmington, Delaware) and another 90 of the variety Deltapine 1646 B2XF (Bayer AG, Leverkusen, North Rhine-Westphalia, Germany) were planted at the test field in Burleson county.

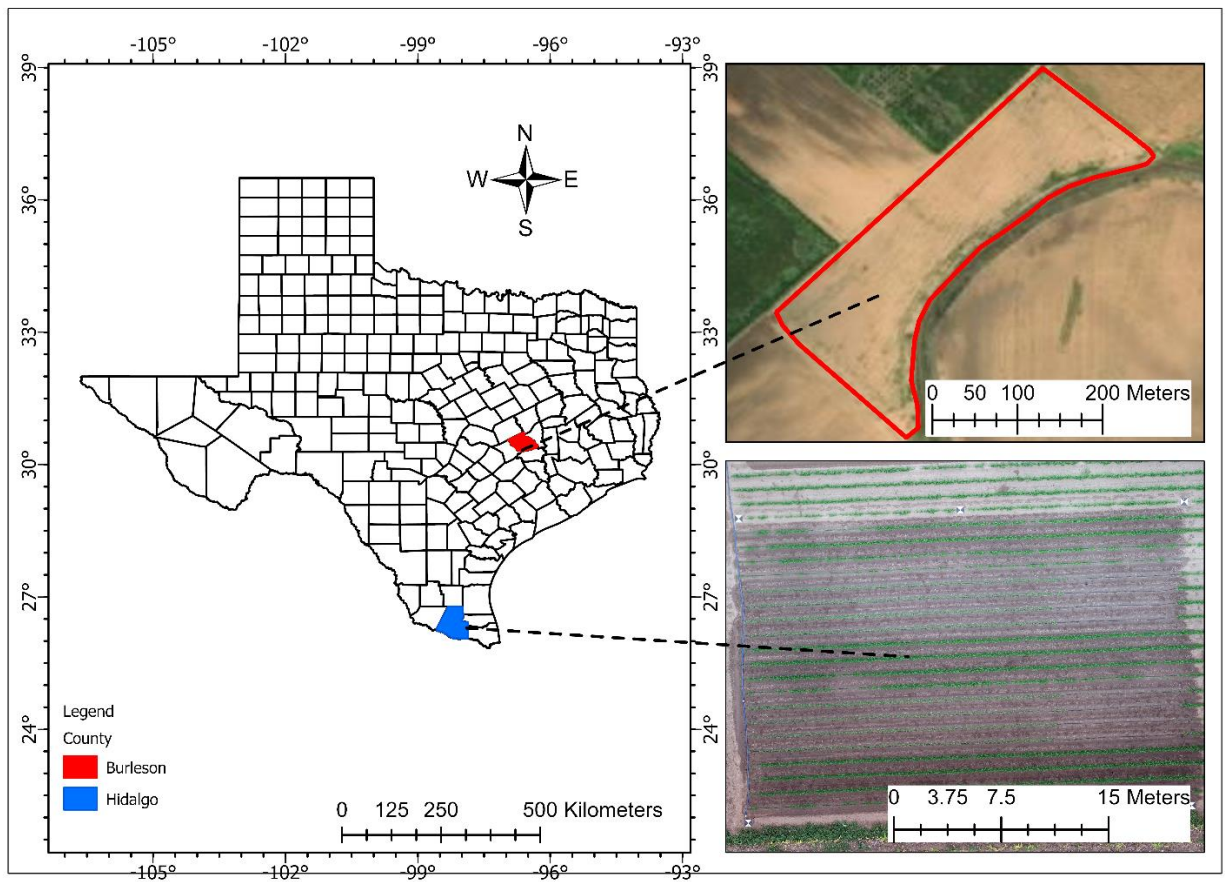

Figure 1: Experiment site of two corn fields located in Burleson and Hidalgo counties.

### 2.2 Image data acquisition

At the experiment site located in Weslaco, Texas, a three band (RGB: red, green, and blue) FC6310 camera (Shenzhen DJI Sciences and Technologies Ltd., Shenzhen, Guangdong, China) integrated on DJI Phantom 4 Pro quadcopter (Shenzhen DJI Sciences and Technologies Ltd., Shenzhen, Guangdong, China) was used to collect aerial imagery from an altitude of 18.3 m (60 ft) above ground level (AGL). The images acquired by the FC6310 camera had a resolution of 5472 x 3648 pixels and a spatial resolution of 0.5 cm/pixel (0.20 in./pixel). Data were collected on April 7, 2020, between 10 a.m. and 2 p.m. Central Standard Time (CST) at 80% sidelap and 75% overlap. The corn plants were at V3 stage when the data were acquired.

At the second site located in Burleson county, RedEdge-MX (AgEagle Aerial Systems Inc, Wichita, Kansas) camera was mounted on a custom UAS (Figure 2) for aerial data collection. The first set of data were collected on May 5, 2021, between 11:00 a.m. and 2:00 p.m. central daylight-saving time (CDT) at an altitude of nearly 4.6 meter (15 feet) above ground level (AGL) when the corn plants were at V6 growth stage. The second set of data were collected on May 14, 2022, between 11:00 a.m. and 2:00 p.m. CDT from an altitude of 4.6-meter (15 feet) AGL when the corn plants had reached the VT stage. The acquired images on both the days had approximate spatial resolution of 0.34 cm/pixel.

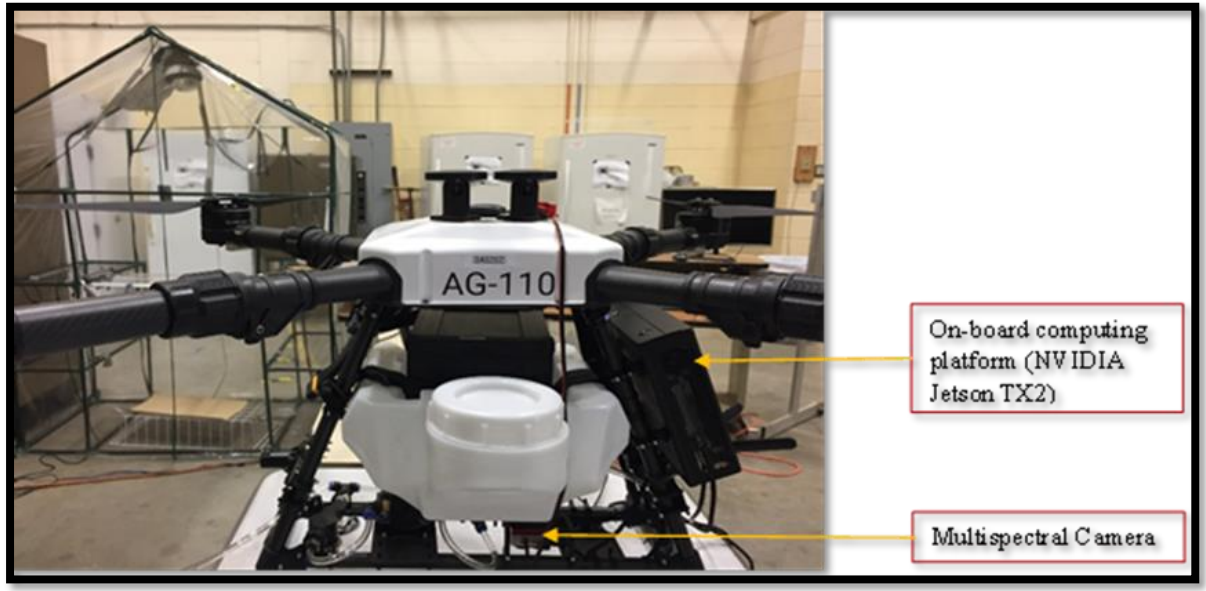

Figure 2: A customized unmanned aircraft system (UAS) with on-board computing platform and MicaSense RedEdge-MX multispectral camera.

**2.3 YOLOv5**

YOLOv5 is the fifth version of YOLO series of object detection algorithm which was released in June of 2020 (Jocher, 2020). Just like its predecessors, it is a single stage detector network which makes it faster compared to other object detection algorithms (Yan et al., 2021; Zhou et al., 2021). The simple architecture of YOLOv5 can be seen in Figure 3 which was generated by a neural network visualization software tool Netron version 4.8.1 (Lutz, 2017). The overall architecture comprises of 25 nodes which are named from model 0 to model 24. Nodes 0 to 9 (i.e., model/0 to model/9) form the backbone network while nodes 10 to 23 (i.e., model/10 to model/23) represent the neck network and the last node i.e., model/24 forms the detection network. The last node comprises of three layers to make detections at three different scales with bounding boxes and confidence scores around the detected objects.

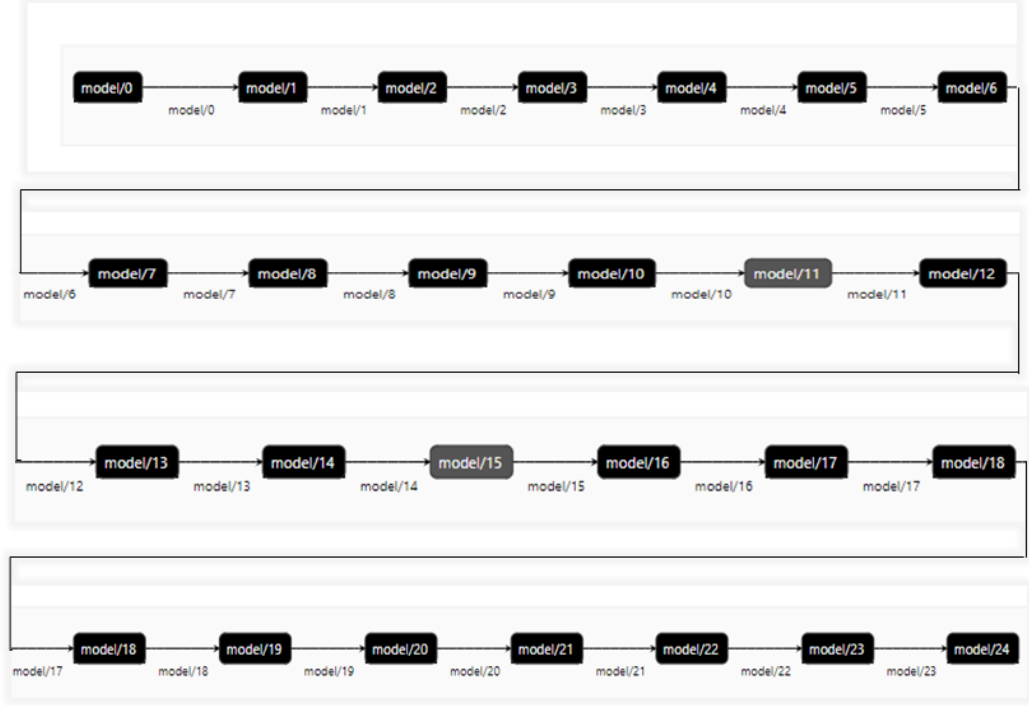

Figure 3: Simplified version of YOLOv5 architecture generated by Netron visualization software tool.

The original YOLOv5 architecture accepts input images of three channels i.e., 3 bands and it comes pretrained on Common Objects in Context (COCO) (Lin et al., 2014) dataset. The original architecture of YOLOv5 was customized to detect a single class of VC plants and accept input images of shape 3 x 416 x 416 for this study.

**2.4 Performance Metrics of YOLOv5**

The performances of each trained YOLOv5 model were assessed by using the metrices accuracy, precision (P), recall (R), mean average precision (mAP) and F1-score as calculated in equations 1, 2, 3, 4 and 5.

$$\text{Accuracy} = \frac{TP+TN}{TP+TN+FP+FN} \qquad (1)$$

$$P = \frac{TP}{TP+FP} \qquad (2)$$

$$R = \frac{TP}{TP+FN} \qquad (3)$$

$$\text{F1-Score} = 2 \times \frac{P \times R}{P+R} \qquad (4)$$

$$\text{mAP} = \frac{1}{n}\sum_{k=1}^{k=N} AP_k \qquad (5)$$

where *TP* is the number of true positives, *TN* is the number of true negatives, *FP* is the number of false positives, *FN* is the number of false negatives, $AP_k$ is the average precision (AP) of class *k* (in our case *N = 1* i.e., VC class), and *n* is the number of thresholds (*n =1* in our case i.e., 0.50 or 50%). This essentially means, the mAP and AP are same for a single class case like in this study. The mAP values were calculated at 50% threshold value for intersection over union (IoU) meaning, all the predicted bounding boxes that resulted in ratios of overlapping areas to the union areas with ground truth bounding greater than 50% were considered and the remaining were discarded (Gan et al., 2021; Padilla et al., 2021; Sharma, 2020; Wu et al., 2021). mAP is a metric based on the area under precision-recall curve (PRC) that is preprocessed to eliminate zig-zag behavior (Padilla et al., 2021). Therefore, in case of class imbalance where there are more instances of one class than the other, mAP is a more reliable and powerful metric to analyze performance of a classifier (Padilla et al., 2021; Saito & Rehmsmeier, 2015).

**2.5 Dataset Preparation for Training Yolov5**

The RGB images collected at experiment plot located in Weslaco, TX, were split into 416 x 416-pixel after which augmentation techniques were applied to images that contained at least a VC plant using Augmentor Python library (Bloice, 2017) that generated a total of 409 images. To accomplish this, *rotate*, *flip_left_right*, *zoom_random* and *flip_top_bottom* operations were used with probability values of 0.80, 0.40, 0.60 and 0.80 respectively as explained by Bloice et al. (2017). Roboflow, a web-based platform was used to annotate ground truth bounding boxes for VC plants (Roboflow, 2022) by first dividing the images in the ratio 16:3:1 for training, validation and test dataset and then exporting them in YOLOv5 PyTorch format. A total of more than 1750 instances for VC class were obtained in the training, validation, and test datasets (Figure 4A).

The five-band multispectral images collected at experiment site located in Burleson county, Texas were radiometrically corrected and then some preprocessing methods were applied using the source code from GitHub (GitHub, Inc., San Francisco, CA, U.S.A.) repository of MicaSense (MicaSense Incorporated, 2022). This generated RGB images of size 1207 x 923 pixels. Among the generated RGB images, only the ones which had VC plants were chosen and then split into size 416 x 416 pixels. This again resulted in many images without VC plants which were again discarded. The resulting images with at least one VC plants were then augmented to generate a total of 387 images at V6 stage and 280 at VT stage. Roboflow was then used to annotate the ground truth bounding boxes for VC plants. Of these, training, validation, and test images were divided in the ratio 16:3:1. A total of more than 800 instances (Figure 4B) of VC class were obtained in the datasets for the V6 stage and 400 for the VT stage (Figure 4C).

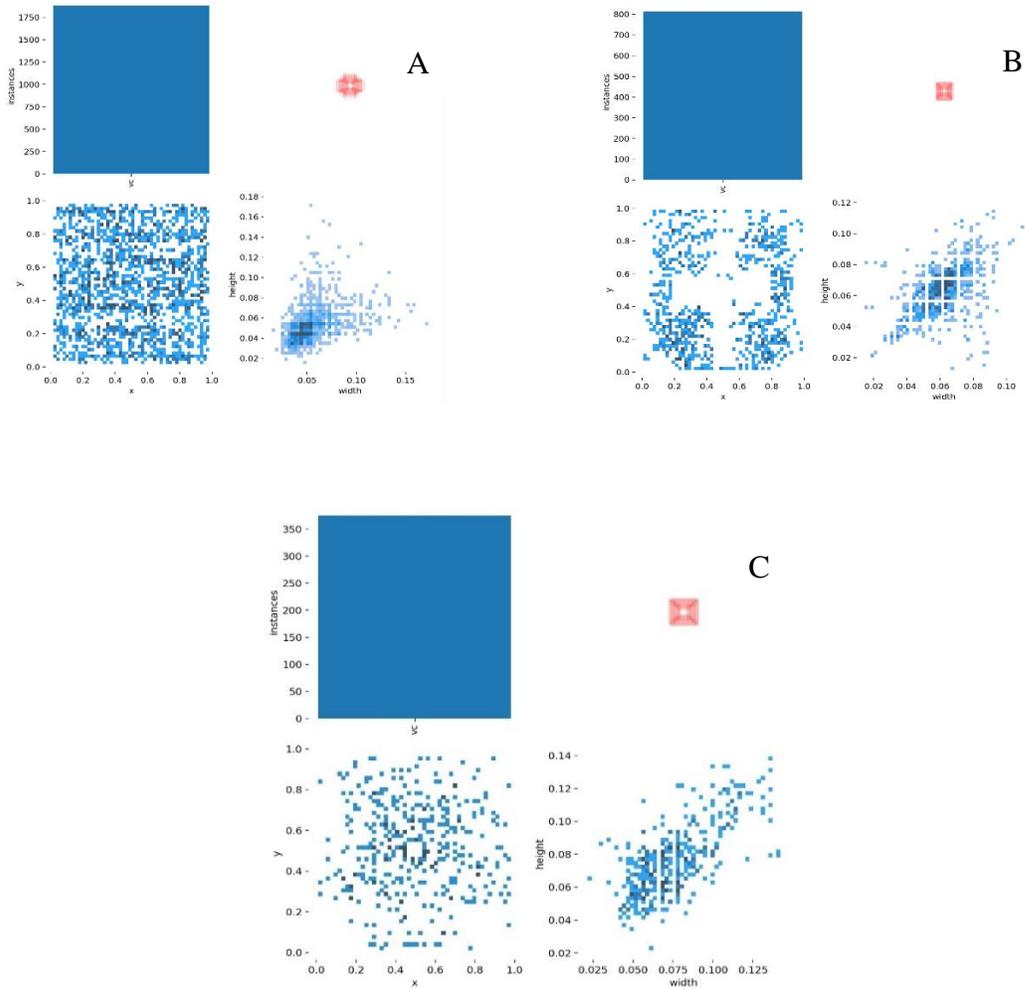

Figure 4: VC class instances in training, validation and test datasets used for the YOLOv5 model at V3 (A), V6 (B) and VT (C) growth stages of corn plants.

**2.6 YOLOv5 Training**

The source code for YOLOv5 was obtained from the GitHub repository of Ultralytics Inc. (Jocher, 2020). All the four variants of YOLOv5 were trained on Tesla P100-PCIE-16 GB (NVIDIA, Santa Clara, CA, U.S.A.) GPU using the Google Colaboratory (Google LLC, Melno Park, CA, U.S.A.) AI platform. Each of them were trained with initial learning rate of 0.01, final learning rate of 0.1, momentum of 0.937, weight decay of 0.0005 for a total of 200  in Figure 5.

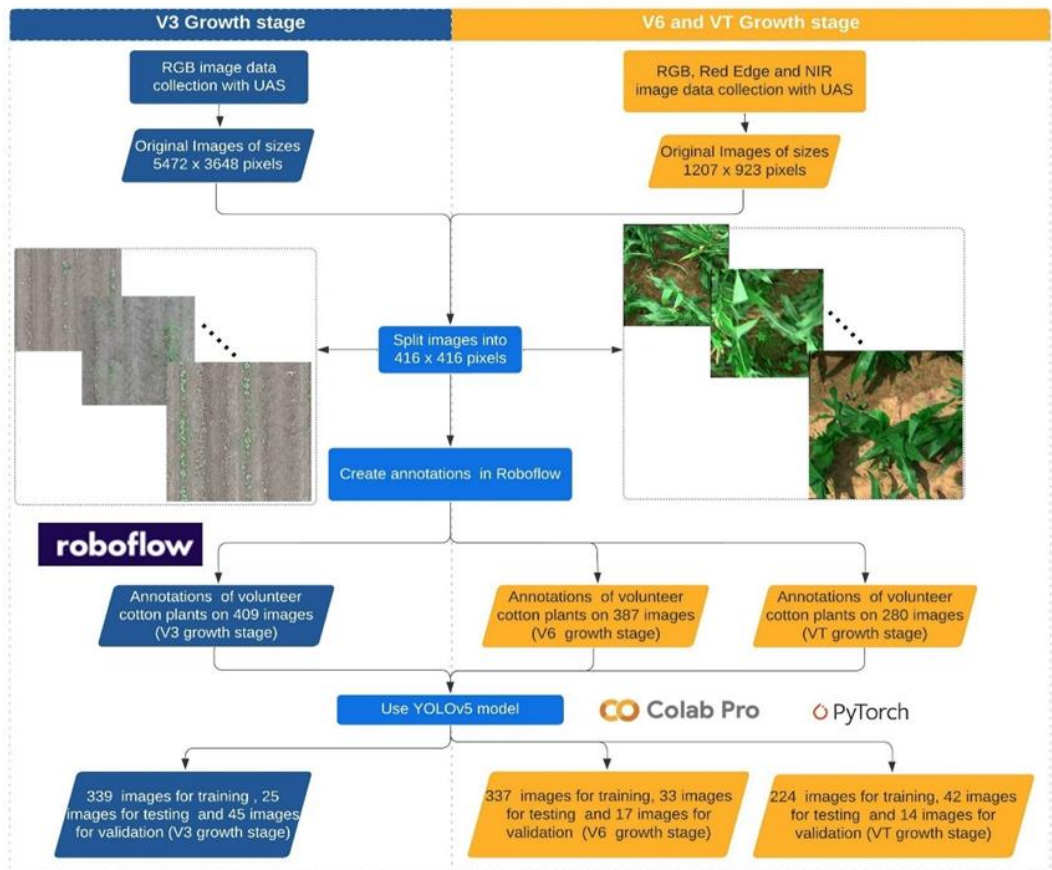

Figure 5: Work flowchart showing entire processes from data acquisition to YOLOv5 training.

## 3.0 Results
### 3.1 VC Detection In a Field With Corn At V3 Growth Stage

The results for different performance metrics that were obtained during the training process using the training and validation datasets are shown in Figure 6.

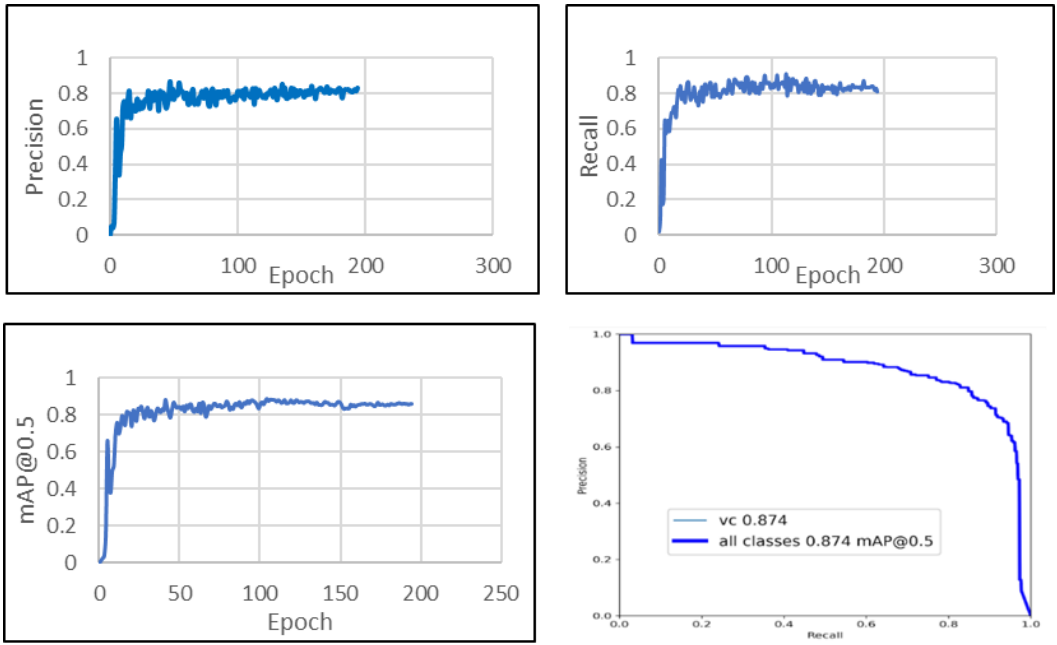

Figure 6: Example graphs showing performance metrics along with PR-curve (PRC) obtained after training YOLOv5s using the dataset belonging to the V3 growth stage of corn plants.

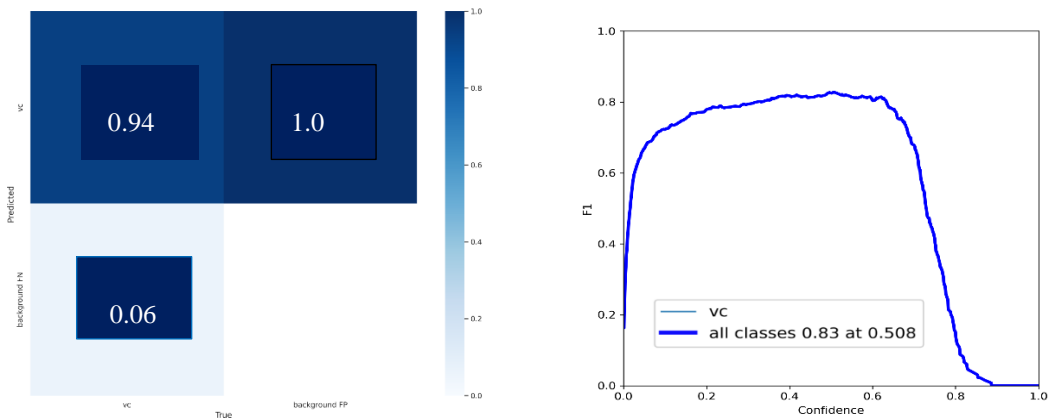

Figure 7: Classification results shown in confusion matrix (left) and F1-score calculated over a range of confidence scores (right) of VC plants after YOLOv5s was trained on dataset belonging to the V3 growth stage of corn plants.

As seen in Figure 6, convergence was achieved within the 200 training iterations by using transfer learning approach in the training process i.e., by starting training from the pretrained weights of YOLOv5 (COCO dataset weights). It was found that the maximum values of precision, recall, mAP, and F1-score reached 87%, 91%, 89% and 83% respectively. Similarly, 94% of the VC plants were correctly classified while 6% were mistaken for background class i.e., either corn plants or weeds. However, none of the instances from the background class were mistaken for VC plants (Figure 7-left). Almost every image had more instances of background class (i.e., corn and weeds) than the VC class, which means class instances were imbalanced. From past studies, it has been found that classifier's performance can be biased towards the majority class, therefore it has been recommended to analyze performance based on the PRC as seen in Figure 6 (Saito & Rehmsmeier, 2015). Some of the detection results of VC plants are shown in Figure 8 where the detected VC plants are enclosed within the red bounding boxes with corresponding confidence scores.

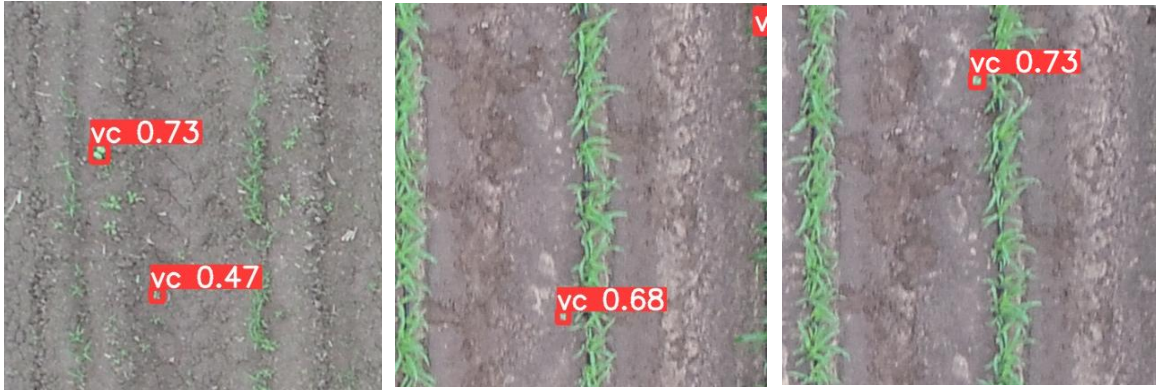

Figure 8: YOLOv5s detection results of VC plants in a field of corn at V3 growth stage.

**3.2 VC Detection In a Field With Corn At V6 Growth Stage**

In Figure 9, the graphs for different performance metrics can be seen that were obtained after training YOLOv5s by using datasets belonging to corn plants at V6 growth stage.

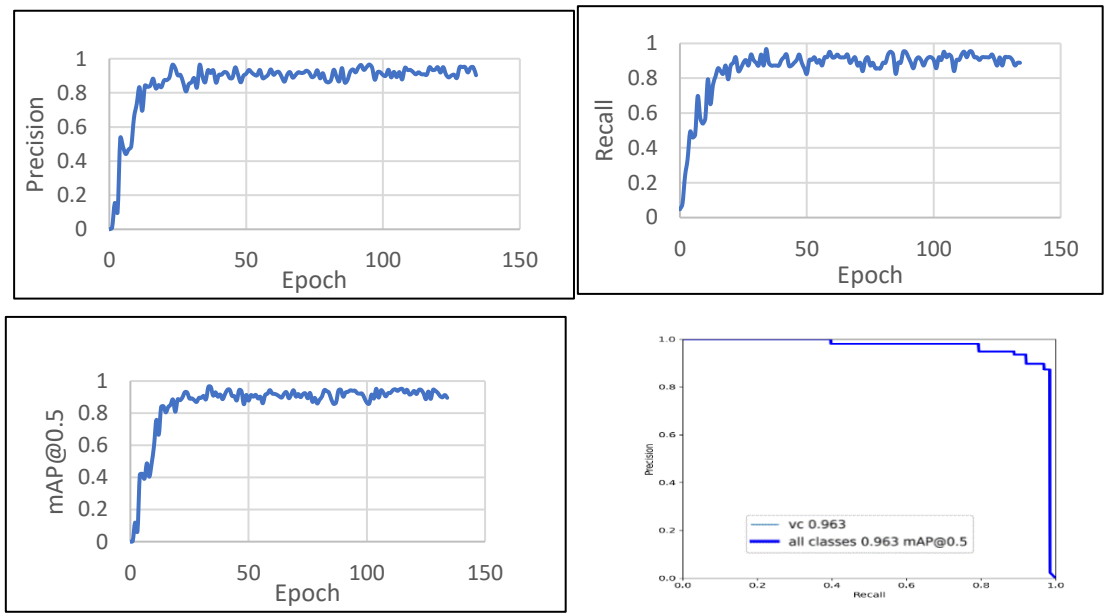

Figure 9: Example graphs showing performance metrics along with PR-curve (PRC) obtained after training YOLOv5s using the dataset belonging to the V6 growth stage of corn plants.

Due to lack of improvement in performance for 100 iterations, there was early stopping of the training around the 130$^{th}$ iteration. This means that convergence was achieved well before the 130$^{th}$ iteration. The maximum values of precision, recall, mAP, and F1-score were found to be 97%, 97%, 96% and 93% respectively. In terms of classification accuracy, 98% of the VC plants were correctly classified while only 2% of them were misclassified as corn or weeds and none of the background class (i.e., corn and weeds) were misclassified as VC plants (Figure 10-left).

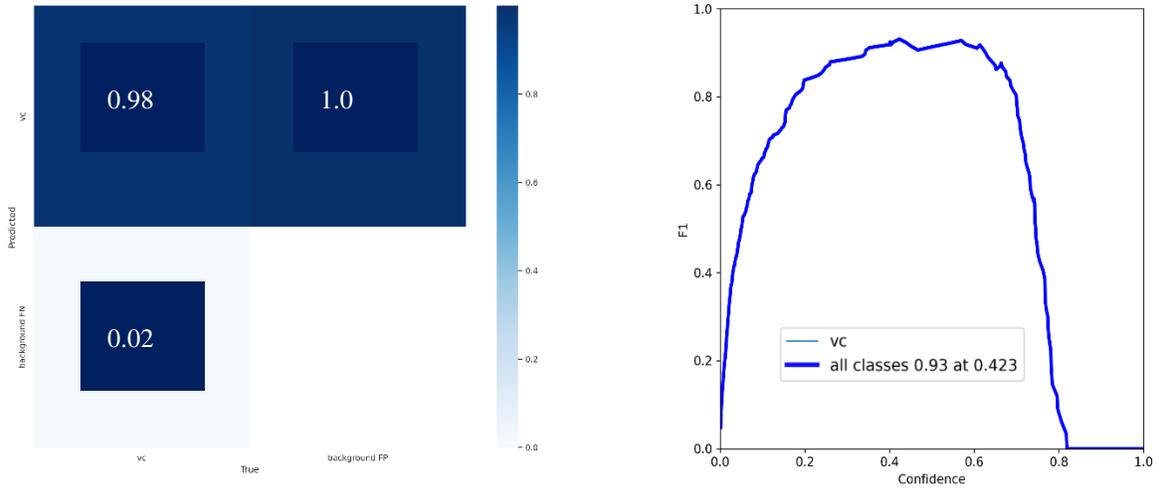

Figure 10: Classification results shown in confusion matrix (left) and F1-score calculated over a range of confidence scores (right) of VC plants after YOLOv5s was trained on dataset belonging to the V6 growth stage of corn plants.

Some of the detection results of VC plants are shown in Figure 11 where the detected VC plants are enclosed within the red bounding boxes with their corresponding confidence scores.

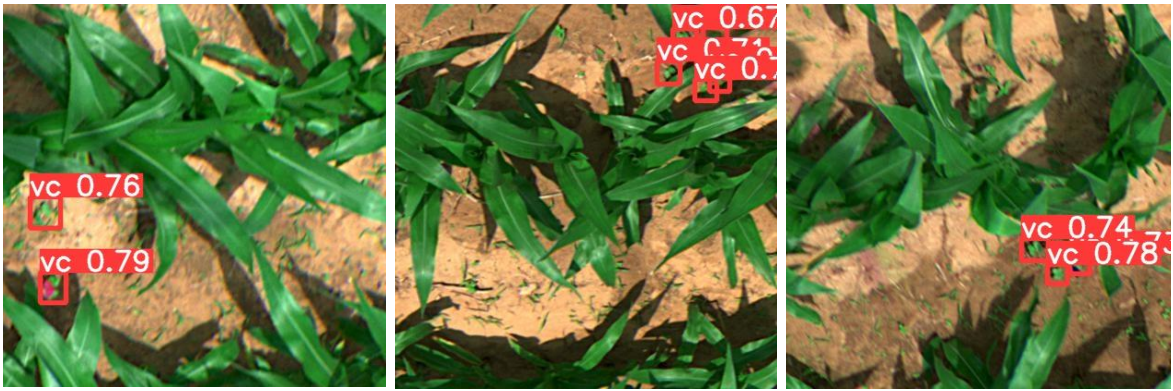

Figure 11: YOLOv5s detection results of VC plants in a field of corn at V6 growth stage.

**3.3 VC Detection In a Field With Corn at VT Growth Stage**

Figure 12 shows the graphs for different performance metrics that were obtained after training YOLOv5s by using datasets belonging to corn plants at VT growth stage.

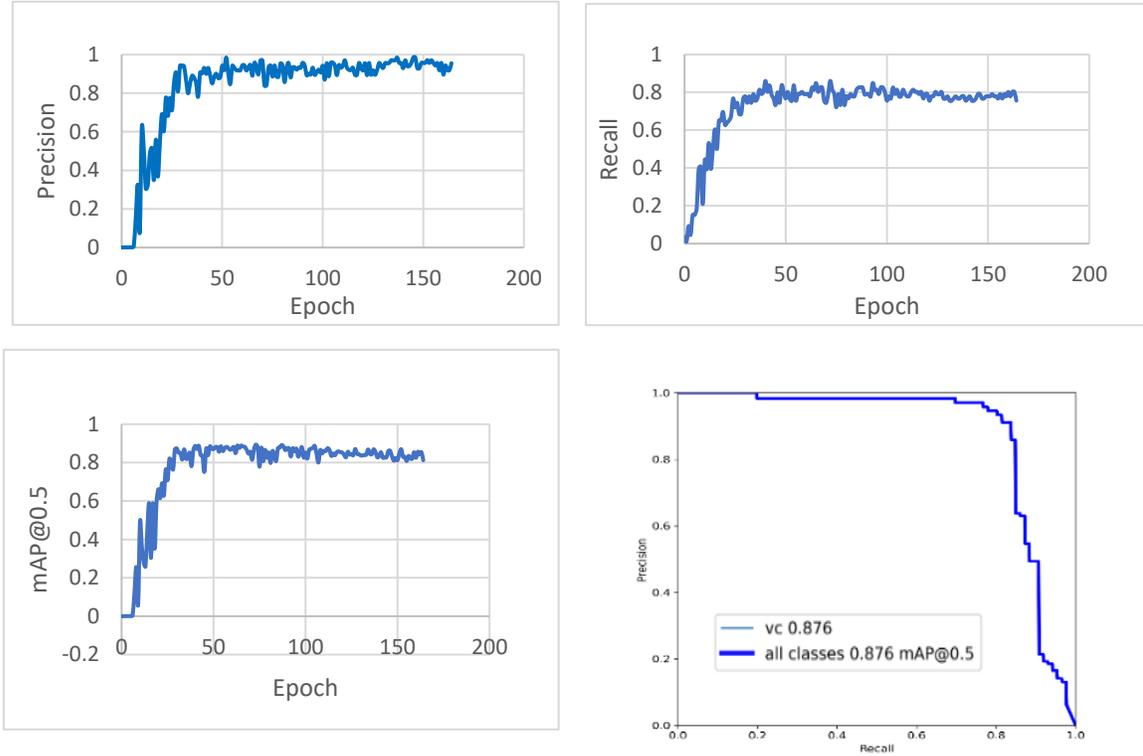

Figure 12: Example graphs showing performance metrics along with PR-curve (PRC) obtained after training YOLOv5s using the dataset belonging to the VT growth stage of corn plants.

As in the V6 case, there was early stopping in the training process around 165[th] iteration for the VT case too as no improvement was observed in the last 100 iterations (Figure 12). This implies that convergence was achieved well before the 165[th] iteration. The maximum values of precision, recall, mAP, and F1-score were found to be 99%, 86%, 89% and 87% respectively. In terms of classification accuracy, 85% of the VC plants were correctly classified while 15% of them were misclassified as corn or weeds and none of the background class (i.e., corn and weeds) were misclassified as VC plants (Figure 13-left).

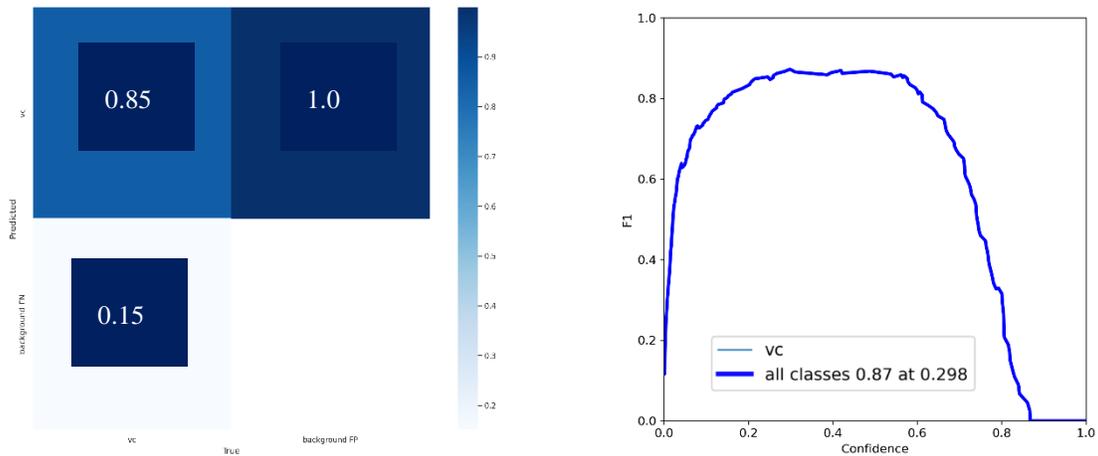

Figure 13: Classification results shown in confusion matrix (left) and F1-score calculated over a range of confidence scores (right) of VC plants after YOLOv5s was trained on dataset belonging to the VT growth stage of corn plants.

Some of the detection results of VC plants are shown in Figure 14 where the detected VC plants are enclosed within the red bounding boxes with their corresponding confidence scores.

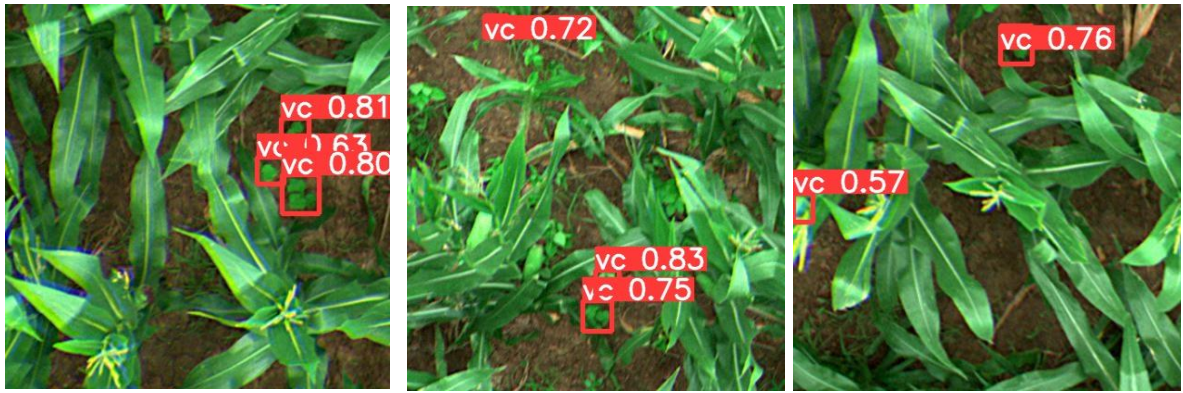

Figure 14: YOLOv5s detection results of VC plants in a field of corn at VT growth stage

**3.4 Comparison of Detection Results At Three Different Growth Stages of Corn With All the Four Variants of Yolov5**

Table 1 shows the comparative results of all the four variants of YOLOv5 in detecting VC plants growing in fields with corn plants at three different growth phases i.e., at V3, V6 and VT.

Table 1: Comparison of performance metrics of four variants of YOLOv5 in detecting VC plants growing in the middle of fields with corn plants at three different growth phases.

| Corn Growth Phase | Classification_Acc YOLOv5 | | | | mAP@50 YOLOv5 | | | | F1-Max YOLOv5 | | | |
|---|---|---|---|---|---|---|---|---|---|---|---|---|
| | s | m | l | x | s | m | l | x | s | m | l | x |
| V3 (early) | 94 | 92 | 90 | 93 | 87.9 | 86.5 | 86.5 | 87.4 | 83 | 82 | 83 | 83 |
| V6 (mid) | 98 | 95 | 92 | 95 | 96.3 | 93.5 | 91.6 | 92.0 | 93 | 92 | 91 | 91 |
| VT (tassel) | 85 | 85 | 91 | 86 | 87.6 | 90.3 | 89.0 | 87.6 | 87 | 88 | 85 | 88 |

It is evident that VC plants were most accurately classified from the V6 growth stage corn than the V3 and VT growth stages by using all the four variants of YOLOv5. In terms of detection accuracy based on mAP at 50% IoU, similar results were observed (Table 1). Between V3 and VT growth stage, VC plants were more accurately classified as well as detected in the V3 than the VT stage.

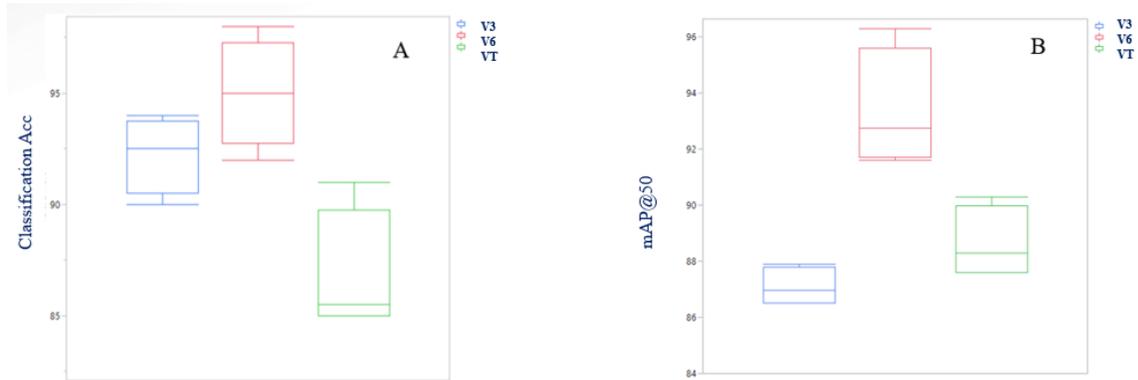

Figure 15: (A) Classification accuracy distribution with all the four variants of YOLOv5 at all the three growth stages of corn plants. (B) mAP distribution with all the four variants of YOLOv5 at all the three growth stages of corn plants.

The average mAP for detecting VC plants in fields with corn plants at V3, V6 and VT growth stages by using all the four variants of YOLOv5 were found to be 87%, 93% and 89% with standard deviations of 0.69, 2.13 and 1.3 respectively. Similarly, the average classification accuracy of VC plants in fields with corn plants at V3, V6 and VT growth stages were found to be 92%, 95% and 87% with standard deviations of 1.7, 2.4 and 2.9 respectively.

## 4.0 Discussion
### 4.1 VC Detection Comparisons

The images acquired at V3 growth stage of corn plants were captured by high resolution camera but at the same time they were captured from an altitude that is four times more than the images that were captured at V6 and V8 growth stages. This resulted in a lower spatial resolution (0.5 cm/pixel) than the images captured at the other two growth stages (0.34 cm/pixel). At this stage, there were not many weeds in the background of images which is why the average classification accuracy of 92% was promising (Figure. 15A). Even though this was an increment of more than 31% than our previous approach (Yadav et al., 2019), the average classification accuracy at the V6 stage was found to be even more at 95% (an increment of nearly 36%). It was expected that the accuracy might be lower at V6 stage as compared to the V3 stage due to larger canopy size of corn plants and relatively more weeds in the background. However, the higher spatial resolution in the imagery by capturing images at lower altitude (4.6 m /15 feet) resulted in a better classification accuracy (Table 1). The other reason for higher accuracy at V6 stage could be due to the fact that the images were preprocessed by using affine transform, unsharp mask and gamma correction (MicaSense Incorporated, 2022). However, these effects were not enough to improve classification accuracy at the VT growth stage (Figure 15A). We assume this was because at this stage the weeds in the background were prominent and caused in more misclassifications than at the V6 stage.

In this study, at all the three growth stages of corn, the instances of VC plants as compared to the instances of background class (i.e., corn with weeds) was always less which means there was imbalance between the positive (VC plants) and negative (background) classes. In such cases, performance of any classifiers are found to be biased towards the majority class and therefore one cannot accurately make inference about the performance (Saito & Rehmsmeier, 2015; Sofaer et al., 2019). In such cases, area under the PRC is found to be more effective measure of the classifier performance (Miao & Zhu, 2021). Therefore, to make fair comparison without results being affected by class instance imbalance, mAP values are used from which the highest detection accuracy was found at the V6 growth stage followed by VT and V2 (Figure 15B). This result was expected because images at V6 and VT stages were of higher spatial resolution as well had undergone preprocessing techniques which were absent in the case of images at V3 stage. The presence of prominence in weeds at VT stage resulted in lower detection accuracy as compared to the V6 stage.

## 5.0 Conclusions and Future Recommendations

There is a tradeoff between detection accuracy and the amount of area surveyed because to survey larger area, images need to be captured from higher altitude like in the case of V3 stage which eventually decreases spatial resolution and a compromise in detection accuracy. However, if 87% mAP is not too low then we recommend of surveying at V3 growth stage for detecting VC plants as it results in detecting them in a larger area. We also expect the detection accuracy to improve and perhaps surpass that of V6 and VT stage when the images can be preprocessed using the techniques described in the GitHub repository of MicaSense (MicaSense Incorporated, 2022). This way, the system will be more practically viable and applicable by the personnel involved in the boll weevil eradication program.

## Acknowledgments

This material was made possible, in part, by Cooperative Agreement AP20PPQS&T00C046 from the United States Department of Agriculture's Animal and Plant Health Inspection Service (APHIS). It may not necessarily express APHIS' views. We would like to extend our sincere thanks to all the reviewers and people involved during field work including Stephen P. Labar, student assistants (Roy Graves, Madison Hodges, Sam Pyka, Reese Rusk, Raul Sebastian, JT Womack, John Marshall, Katelyn Meszaros, Lane Fisher, and Reagan Smith), Dr. Thiago Marconi, and Uriel Cholula.